\documentclass{article}
\PassOptionsToPackage{numbers, compress}{natbib}

\usepackage[preprint]{neurips_2026}

\usepackage[utf8]{inputenc} 
\usepackage[T1]{fontenc}    
\usepackage{hyperref}       
\usepackage{url}            
\usepackage{booktabs}       
\usepackage{amsfonts}       
\usepackage{nicefrac}       
\usepackage{microtype}      
\usepackage{xcolor}         
\usepackage{enumitem}       

\usepackage{graphicx}
\usepackage{amsmath}
\usepackage{amsthm}
\usepackage{algorithm}
\usepackage{algorithmic}
\usepackage{subcaption}
\usepackage{listings}

\newtheorem{definition}{Definition}
\usepackage{amsfonts}
\usepackage{amssymb}
\usepackage{booktabs} 
\usepackage{multirow}
\usepackage{threeparttable}

\newtheorem{example}{Example}
\newtheorem{theorem}{Theorem}

\title{Verifiable LLM-Guided Focal SMT Solving for Quantified Arrays}

\author{%
    Kunhang Lv\textsuperscript{1,3} \quad
    Rui Han\textsuperscript{2,3} \quad
    Fuqi Jia\textsuperscript{2,3}\thanks{Corresponding authors.} \quad
    Yuhang Dong\textsuperscript{1,3} \quad
    Feifei Ma\textsuperscript{1,3}\footnotemark[1] \quad
    Jian Zhang\textsuperscript{2,3} \\
    \normalfont
    \textsuperscript{1}Laboratory of Parallel Software and Computational Science, ISCAS, Beijing, China\\
    \textsuperscript{2}SKLCS and Key Laboratory of System Software, ISCAS, Beijing, China\\
    \textsuperscript{3}University of Chinese Academy of Sciences, Beijing, China\\
    \texttt{\{lvkh, jiafq, maff\}@ios.ac.cn}
}

\begin{document}

\maketitle

\begin{abstract}
  Large language models can often identify useful semantic structure in formal artifacts, but their outputs cannot be trusted as proofs or solver decisions. We study whether LLMs can instead guide formal reasoning systems through a narrow, solver-verifiable interface. We instantiate this idea in quantified array SMT solving, where modern instantiation-based techniques can fail when key ground terms and equalities are hidden by auxiliary variables, axiom guards, and nested array terms.
  We observe that many instances are largely driven by a small set of focal ground terms or constraints, similar to the backdoor sets in SAT solving. Building on this, we present \textsc{LinguaArray}, a framework where an LLM proposes a small \emph{Semantic Focus Set} (ground terms/constraints), while a conventional SMT solver remains the sole proof engine and certifies all results. \textsc{LinguaArray} uses these LLM-suggested, solver-validated finite sets to significantly enhance solver capability, solving additional satisfiable and unsatisfiable instances that the backend alone fails to solve. It interacts within a finite time and then falls back to the backend solver to mitigate potential regressions.
  Evaluation on 1015 SMT-LIB instances across five logics under 1200s timeout (including LLM latency) shows that \textsc{LinguaArray} solves up to 69.1\% of instances and increases solved-instance counts by 105\%--397\% over the corresponding backends.
\end{abstract}

\section{Introduction}

Large language models have shown promise in reasoning over structured symbolic inputs, but formal reasoning tasks require correctness guarantees beyond what unverified LLM outputs provide. This motivates a restricted form of LLM assistance: using LLMs to propose semantic guidance while leaving all correctness-critical decisions to a trusted symbolic solver. We study this question in the context of Satisfiability Modulo Theories (SMT)~\cite{DBLP:series/txtcs/KroeningS16,DBLP:series/faia/BarrettSST21}, a key computational foundation for software verification, program analysis, model checking, and automated test generation~\cite{DBLP:journals/jar/BeyerDW18,DBLP:conf/cav/GavrilenkoLFHM19,DBLP:conf/kbse/CordeiroFM09,DBLP:conf/nfm/PeleskaVL11}. Among them, quantified arrays are widely used to express loop invariants, memory safety conditions, and relational specifications over data structures, but remain challenging for modern solvers.
The combination of arrays and quantifiers is generally complex and undecidable~\cite{bradley2006s}, which is particularly challenging for SMT solvers.

Modern SMT solvers primarily handle quantified formulas through quantifier instantiation, such as trigger-based E-matching~\cite{de2007efficient,DBLP:conf/tacas/MouraB08} and model-based quantifier instantiation (MBQI)~\cite{ge2009complete}.
Selecting effective triggers and diagnosing missing instantiations remain major bottlenecks~\cite{becker2019axiom}, and even recent refinements~\cite{reynolds2015counterexample,niemetz2021syntax,reynolds2018revisiting} still depend on identifying a small set of proof-relevant ground terms early.
For array theory, solvers typically use the read-over-write and extensional axiom for simplification and reasoning~\cite{DBLP:conf/lics/StumpBDL01,brummayer2008lemmas}. However, when quantifiers are present and combined with complex operations such as these, key indexes and relationships are often obscured, which can lead to insufficient instantiation~\cite{reynolds2013quantifier,becker2019axiom}.

Consider a simple quantified formula over arrays:
$$
\forall i.\ (0 \le i < n \Rightarrow A[t \mapsto x][i] \ge 0).
$$
This uses the write operation on $A$ and verifies under the universal quantifier. It's obvious we need to locate the assignment $x$ at position $t$ to determine satisfiability; the solver must ultimately instantiate the universal quantifier at the witness index $i=t$.
In realistic benchmarks, such witness terms are rarely presented as simple variables: they are buried under auxiliary definitions, and can even be \emph{array-valued} terms when quantifiers range over arrays (cf.~Example~\ref{ex:motiv}).
As a result, it's hard for the solvers to find the exact instantiation candidates, leading to timeouts or abandonment.

Therefore, the difficulty of quantified array solving does not lie in the inference rules of arrays themselves.
Instead, a few decisive \emph{semantic focusing} choices are often sufficient to render the problem readily tractable for the solver.
However, traditional solvers rely solely on fixed syntactic triggers. 
This prevents them from exploiting the diverse and deep semantic information in real-world formulas.
Our insight is that this missing capability cannot be systematically recovered through trigger engineering or hard-coded rules; instead, it can be supplied by a large language model (LLM) and the solving process can be safely accelerated by limiting it to a focus object that is verifiable by the solver.

To surface such focal points, we propose \textsc{LinguaArray}, a framework that leverages LLMs to provide semantic guidance for quantified array solving. The key idea is that LLM identifies and proposes relevant focal points, which we call the semantic focus set, while using an SMT solver as the backend engine to validate all candidates and prove the final result. We realize this idea with a two-branch architecture. 
The UNSAT branch uses LLM-proposed \emph{Instantiation Focus Sets}, i.e., a small collection of candidate ground instantiation tuples for universally quantified variables, to steer quantifier instantiation. 
The SAT branch uses \emph{Constraint Focus Sets}, i.e., salient ground constraints, to guide model construction. 
Each branch constructs a derived formula, and once its satisfiability state is determined by the backend solver, yields a certified answer to the original problem. 
Since all conclusions are verified by the solver, soundness is preserved regardless of suggestion quality. To preserve baseline coverage, the LLM-guided interaction runs under a bounded budget and the framework falls back to the backend solver on the original formula under the remaining timeout.

We evaluate \textsc{LinguaArray} on 1015 quantified array instances from SMT-LIB across five logics (ALIA, ABV, ABVFP, ABVFPLRA, ANIA) with a 1200-second end-to-end timeout per instance (including LLM latency). 
With \textsc{Z3} as the backend solver, \textsc{LinguaArray} solves 701 instances (69.1\%) versus 141 instances (13.9\%) for \textsc{Z3} alone; with \textsc{CVC5} it solves 659 instances (64.9\%) versus 321 instances (31.6\%) for \textsc{CVC5} alone.  Overall, \textsc{LinguaArray} substantially improves solver effectiveness, raising the overall solve rate from 32\% to roughly 70\%.



\paragraph{Contributions.} This paper makes three contributions:
\begin{itemize}[left=0em]
    \item We introduce \emph{Semantic Focus Sets} as a principled interface between LLMs and SMT solvers for quantified array reasoning, and implement this idea in the \textsc{LinguaArray} framework with formal soundness guarantees and graceful degradation.
    \item We design a two-branch strategy, focused instantiation for UNSAT reasoning and focused strengthening for SAT reasoning, that exploits LLM-proposed focal points while preserving solver-certified correctness.
    \item We conduct extensive experiments on SMT benchmarks, demonstrating 105\%-397\% improvements over SOTA solvers, with evaluations confirming the effectiveness of our approach across different LLMs and configurations.
\end{itemize}

\section{Preliminaries}

\subsection{Quantified Arrays in SMT}
\paragraph{SMT.}
Satisfiability Modulo Theories extends propositional satisfiability (SAT) by deciding the satisfiability of first-order formulas with respect to background theories $\mathcal{T}$. A formula $\varphi$ is \emph{$\mathcal{T}$-satisfiable} if there exists a $\mathcal{T}$-model $\mathcal{M}$ such that $\mathcal{M} \models \varphi$. Common theories include arithmetic, bit-vectors, and arrays. Modern SMT solvers such as \textsc{Z3}~\cite{DBLP:conf/tacas/MouraB08} and \textsc{CVC5}~\cite{DBLP:conf/tacas/BarbosaBBKLMMMN22} are typically based on the DPLL(T) framework~\cite{nieuwenhuis2006solving}.
Our work focuses on the challenging combination of quantifiers and arrays theories.

\paragraph{Arrays.}

Arrays are a fundamental data structure in SMT for modeling memory and mappings~\cite{brummayer2008lemmas}. The SMT-LIB theory \texttt{ArraysEx} is parameterized by two sorts: index sort $I$ and element sort $V$. We denote the array sort as $Array(I,V)$.
The theory provides two operations:
\begin{align}
select &: Array(I,V) \times I \to V, \\
store &: Array(I,V) \times I \times V \to Array(I,V),
\end{align}
where $select$ reads an element at an index and $store$ produces an updated array. Their semantics are defined by the following universally quantified axioms ($\forall a,i,j,v$):
\begin{align}
\text{(RW1)} \quad & select(store(a,i,v),i) = v \label{eq:rw1}\\
\text{(RW2)} \quad & i \neq j \Rightarrow select(store(a,i,v),j) = select(a,j) \label{eq:rw2}
\end{align}
and the extensionality axiom stating that arrays are equal if they agree at all indices:
\begin{equation}
\text{(Ex)} \quad \Big(\forall i{:}I.\, select(a,i) = select(b,i)\Big) \Rightarrow a = b \label{eq:ext}
\end{equation}

\paragraph{Quantified arrays.}
Quantifiers are often needed to express array-wide invariants, such as properties that range over indices. We denote the extension of theory $\mathcal{T}$ with arrays as $\mathcal{T}_A$, and its quantified extension as $\mathrm{Q}(\mathcal{T}_A)$.

\begin{definition}[Quantified Array Theory]
The logic $\mathrm{Q}(\mathcal{T}_A)$ extends quantifier-free array formulas with quantifiers over index and element variables. Formulas in $\mathrm{Q}(\mathcal{T}_A)$ are defined inductively:
\begin{equation}
\varphi ::= \text{QF-formula} \mid (\forall \vec{x}{:}\vec{S}.\, \varphi) \mid (\exists \vec{x}{:}\vec{S}.\, \varphi)
\end{equation}
where $\vec{x} = (x_1,\ldots,x_n)$ are distinct variables with sorts $\vec{S} = (S_1,\ldots,S_n)$ from $\mathcal{T}_A$, and QF-formula denotes quantifier-free formulas over $\mathcal{T}_A$.
\end{definition}


\begin{example}
Consider the SMT-LIB style constraint mentioned in the introduction:
\[
\begin{aligned}
&(0 \le t < n)\ \wedge\ (x < 0)\ \wedge\; \forall i:I.\ \bigl(0 \le i < n \Rightarrow
select(store(a,t,x), i) \ge 0\bigr).
\end{aligned}
\]

It is UNSAT, but the contradiction is exposed only after instantiating the quantifier at the witness index $i=t$ and then (RW1) gives $select(store(a,t,x),t)=x$.
\end{example}

\subsection{Solving Quantified Array Formulas}
Quantified array SMT formulas combine array reasoning with quantifier instantiation. For the quantifier-free array theory, axioms~\eqref{eq:rw1}--\eqref{eq:rw2} are typically applied as local rewrite rules to simplify $select(store(\cdots))$ terms. 
For array disequalities $a \neq b$, solvers instantiate the extensionality axiom~\eqref{eq:ext} by introducing a witness index $k$ with the lemma $select(a,k) \neq select(b,k)$.

Quantified formulas are further challenging, and there is generally no complete decision procedure that works across theories. In practice, modern SMT solvers rely on \emph{quantifier instantiation} to reduce quantified formulas to ground instances~\cite{reynolds2013quantifier}.

\paragraph{E-matching.}
Trigger-based pattern matching (E-matching) ~\cite{DBLP:conf/tacas/MouraB08} is a quantifier-instantiation technique widely used in SMT solvers. Given a quantified formula $\forall\vec{x}.\,\psi(\vec{x})$, the solver selects one or more \emph{triggers} from $\psi$. During search, it maintains an \emph{E-graph}, which is a congruence-closure data structure over ground terms. When a term $t$ matches a trigger $p$ modulo the current equalities $E$ (i.e., $E \models t \simeq \theta(p)$ for some substitution $\theta$), the solver adds the instance $\psi(\theta(\vec{x}))$ to the constraint set.

\paragraph{MBQI.}
Model-Based Quantifier Instantiation (MBQI)~\cite{ge2009complete} constructs a candidate model $M$ for the current ground constraints and checks each quantified clause against $M$. If model $M$ violates a quantified clause, MBQI extracts a counterexample witness and generates a corresponding ground instance.



\subsection{LLMs for Semantic Guidance}

Large Language Models are neural networks based on the Transformer architecture~\cite{vaswani2017attention}, trained on vast text corpora to predict the next token in a sequence. Mainstream LLMs like Gemini~\cite{team2023gemini} and DeepSeek~\cite{liu2024deepseek} exhibit strong reasoning capabilities through scale and self-attention mechanisms that capture long-range dependencies, developing internal representations that encode semantic relationships and domain knowledge~\cite{lewkowycz2022solving,wei2022emergent}.

\paragraph{Prompt Engineering.}
LLM performance critically depends on \emph{prompt design}---the systematic construction of inputs to get desired behaviors~\cite{liu2023pre,sahoo2024systematic}. Advanced techniques include chain-of-thought prompting~\cite{wei2022chain} for explicit reasoning steps and role-based prompting~\cite{kong2024better} to invoke domain expertise.

\paragraph{Integration with Formal Systems.} The complementary nature of LLMs and formal systems enables hybrid architectures where LLMs provide semantic insights and hypothesis generation, while symbolic systems ensure logical correctness and verification. 

\section{Methodology}\label{sec:method}

\begin{figure}[!ht] 
  \centering
  \includegraphics[width=1\textwidth]{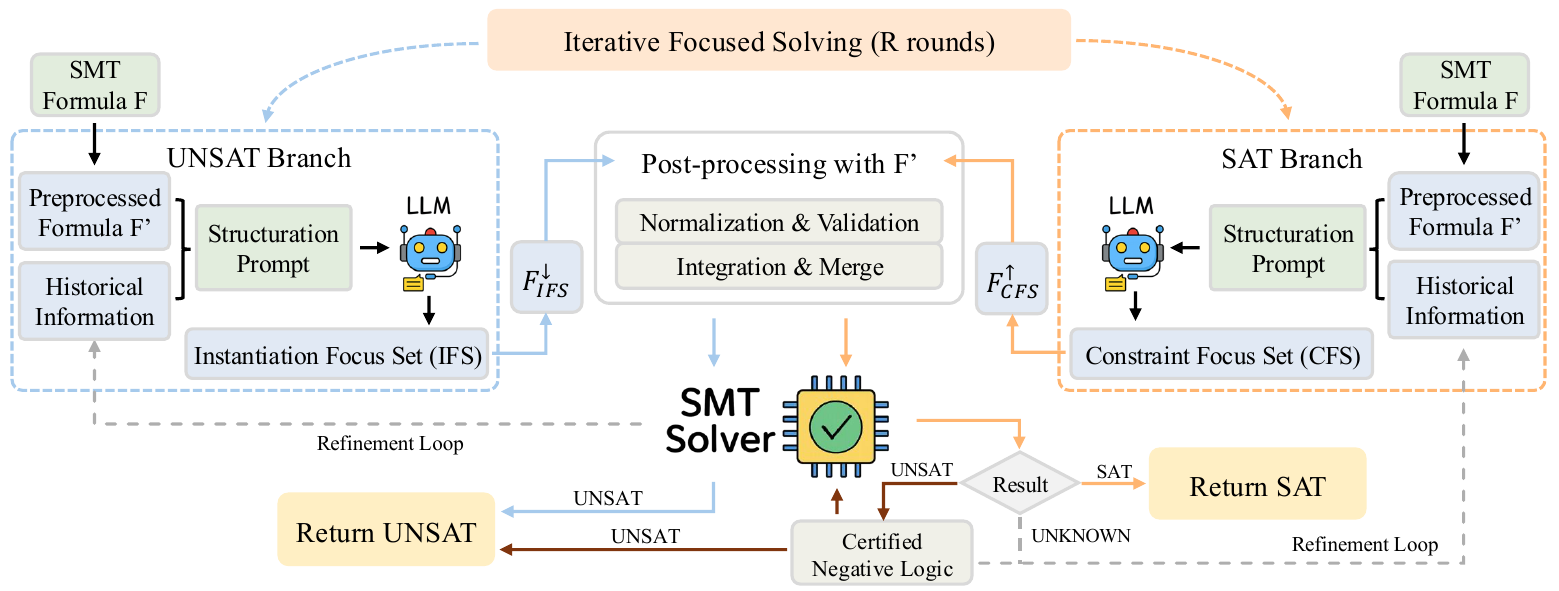}
  \caption{Overview of LinguaArray.}
  \label{fig:mainalg}
\end{figure}

\subsection{Motivating Example}\label{sec:example}

\begin{example}\label{ex:motiv}
Let $a,b : \mathbb{Z} \to (\mathbb{Z} \to \mathbb{Z})$ be 2D arrays and let $i,j,k \in \mathbb{Z}$.
We write $a[t][u] \triangleq \mathsf{select}(\mathsf{select}(a,t),u)$ and $a[i\mapsto x] \triangleq \mathsf{store}(a,i,x)$.
Consider the conjunction:
\begin{equation}
\label{eq:example}
\begin{aligned}
F ={}& \forall x:\text{Array}(\mathbb{Z},\mathbb{Z}).\; (a[i\mapsto x])[j][0] = k\; \wedge\; \exists y:\text{Array}(\mathbb{Z},\mathbb{Z}).\; (b[i\mapsto y])[j][0] \neq k.
\end{aligned}
\end{equation}
\end{example}

This constraint involves universal quantifiers for array-type variables, where the domain of $\text{Array}(\mathbb{Z},\mathbb{Z})$ is infinite, and the solver cannot find the corresponding instantiation position efficiently. Therefore, traditional solvers cannot simplify and give up solving the instance. However, we can find the semantic cases from the reasoning level: if $i\neq j$, then $(a[i\mapsto x])[j][0]$ simplifies to $a[j][0]$, which is independent of $x$; if $i=j$, then the universal quantifier becomes $\forall x.\; x[0]=k$, which is obviously unsatisfiable. Therefore, we find that when $i\neq j$, it is easy to deduce that the original problem is satisfied.

This reflects our core principle: LLM proposes semantically motivated candidates (e.g., $i\neq j$), and the solver verifies them through its standard procedure. Next, we will introduce in detail the framework for implementing this workflow.

\subsection{Framework Overview}\label{sec:overview}

Figure~\ref{fig:mainalg} presents \textsc{LinguaArray}, a hybrid framework where an LLM provides semantic candidates and a standard SMT solver $\mathcal{S}$ serves as the proof engine.

\paragraph{Preprocessing.}
Before querying the LLM, we preprocess the input formula $F$ to facilitate robust generation: (1)~Parse the input of the SMT-LIB file into negation normal form (NNF) to make quantifier polarity explicit. (2)~Simplification: constant propagation, elimination of trivial subformulas, and renaming verbose identifiers to short canonical forms via a reversible mapping $\Sigma$. The resulting $F'$ is equisatisfiable with $F$ while offering a concise representation for the LLM.

\paragraph{Semantic Focus Sets.}
We propose the concept of \emph{Semantic Focus Sets}, intended as an interface between untrusted LLMs and trusted solvers, thereby accelerating the solver.

\begin{definition}[Semantic Focus Sets]\label{def:sfs}
A Semantic Focus Set is a finite structured set proposed by the LLM that identifies where the solver should focus its reasoning. We define two variants:
\begin{itemize}[left=0em]
\item \textbf{Instantiation Focus Set (IFS)}: A finite set of ground instantiation terms. Given a positive occurrence of $\forall x.\, \varphi(x)$ in NNF, each $t \in \mathit{IFS}$ generates the ground consequence $\varphi[x \mapsto t]$. The IFS induces an over-approximation by restricting universal quantifiers to a finite domain.
\item \textbf{Constraint Focus Set (CFS)}: A finite set $\{\psi_1, \ldots, \psi_n\}$ where each $\psi_i$ is a quantifier-free ground formula over the vocabulary of $F'$. The CFS induces an under-approximation of the search space by imposing auxiliary assumptions to guide model construction.
\end{itemize}
\end{definition}

The IFS is generated from terms containing universal quantifiers in the NNF formula $F'$. LLM provides key insights; it may be a complex term (e.g., $t=a_1+a_2$), but only accepts terms that satisfy the corresponding quantifier declaration and can be constructed from symbols in $F'$.

Each constraint in the CFS acts as an auxiliary assumption that helps guide solving (e.g. $i \neq j$ or array cell assignments) and it only accepts constraints that can be constructed from symbols defined in $F'$.

This abstraction provides two key guarantees: (1) the LLM can affect the solver only by generating validated focus sets, and (2) every result is certified by the solver independently of the LLM.

\subsection{Dual-Branch Reasoning}\label{sec:reasoning}

Building on semantic focus sets, we employ two complementary strategies to decide satisfiability.

\paragraph{UNSAT Branch.}
This branch attempts to prove formula unsatisfiability by extracting finitely many ground consequences of universal quantifiers using terms from the IFS.

\begin{definition}[Focused Instantiation]\label{def:finst}
Given an IFS, the focused instantiation $F^{\downarrow}_{\mathit{IFS}}$ is obtained from $F'$ by replacing positive occurrences of universal subformulas $\forall \vec{x}.\,\varphi(\vec{x})$ in NNF with the finite conjunction $\bigwedge_{\vec{t} \in \mathit{IFS}^{|\vec{x}|}} \varphi[\vec{x} \mapsto \vec{t}]$.
\end{definition}

For example, given the positive universal quantifier $\forall x.\, (a[i\mapsto x])[j][0] = k$ and $\mathit{IFS}=\{b[j]\}$, this operation produces a single ground constraint: $(a[i\mapsto b[j]])[j][0] = k$. The resulting check focuses the solver on ground consequences selected by the IFS, avoiding reliance on heuristic triggers for those instances.

\begin{theorem}[UNSAT Branch Soundness]\label{thm:unsat}
$F' \Rightarrow F^{\downarrow}_{\mathit{IFS}}$. Hence, if $\mathcal{S}(F^{\downarrow}_{\mathit{IFS}}) = \mathsf{unsat}$, then $F$ is unsatisfiable.
\end{theorem}

\begin{proof}
For any ground $\vec{t}$, $\forall \vec{x}.\,\varphi \Rightarrow \varphi[\vec{x} \mapsto \vec{t}]$; finite conjunction preserves implication. Since the replacement is applied only to positive NNF occurrences, monotonicity of the surrounding context preserves the implication. All remaining construction steps are required to be implication-preserving, so $F' \Rightarrow F^{\downarrow}_{\mathit{IFS}}$, and unsatisfiability of the consequent implies unsatisfiability of the antecedent.
\end{proof}

\paragraph{SAT Branch.}
This branch attempts to prove satisfiability by imposing CFS constraints as additional assumptions, restricting search to an LLM-suggested semantic region.

\begin{definition}[Focused Strengthening]\label{def:fstr}
Given a CFS, the focused strengthening is $F^{\uparrow}_{\mathit{CFS}} = F' \wedge \bigwedge_{c \in \mathit{CFS}} c$.
\end{definition}

For instance, suggesting a constraint like $i \neq j$ or $k = 5$ prunes the vast model search space, guiding the solver toward a specific semantic region.

\begin{theorem}[SAT Branch Soundness]\label{thm:sat}
$F^{\uparrow}_{\mathit{CFS}} \Rightarrow F'$. Hence, if $\mathcal{S}(F^{\uparrow}_{\mathit{CFS}}) = \mathsf{sat}$, then $F$ is satisfiable.
\end{theorem}

\begin{proof}
Since $F^{\uparrow}_{\mathit{CFS}} = F' \wedge \bigwedge_{c \in \mathit{CFS}} c$, any model satisfying $F^{\uparrow}_{\mathit{CFS}}$ must satisfy $F'$, and therefore satisfies the original formula $F$.
\end{proof}

\paragraph{Certifying UNSAT under strengthening.}
If $F^{\uparrow}_{\mathit{CFS}}$ is $\mathsf{unsat}$, the result may be spurious since CFS constraints might exclude valid models. We certify this result using \emph{implication checking}: for each $c \in \mathit{CFS}$, we verify if $F' \Rightarrow c$ by checking $\mathcal{S}(F' \wedge \neg c) = \mathsf{unsat}$. If all checks pass, then $F' \Rightarrow c$ for all $c$, hence $F' \equiv F^{\uparrow}_{\mathit{CFS}}$ and UNSAT transfers. Notably, if any check yields SAT, we immediately obtain a satisfying model of $F'$.

\subsection{Feedback-Driven Refinement}\label{sec:feedback}

\textsc{LinguaArray} iterates for a bounded budget $R$, in each round executing both the UNSAT and SAT branches and refining their under- and over-approximations based on fresh solver feedback.

\paragraph{Execution cycle.}
At iteration $r$, the system explores both branches, and the LLM generates a proposal conditioned on the cumulative feedback state $\mathit{fb}$.

In the UNSAT branch, the LLM proposes an IFS. When $\mathcal{S}(F^{\downarrow}_{\mathit{IFS}}) = \mathsf{sat}$, the current instantiation is insufficient to derive a contradiction. We extract feedback from the model, specifically array indices that received arbitrary assignments and guards that remained unconstrained, and query the LLM for additional instantiation terms in subsequent rounds.

In the SAT branch, the LLM proposes a CFS. When $\mathcal{S}(F^{\uparrow}_{\mathit{CFS}})$ times out or certification fails, the current CFS is ineffective. We record the failed constraints as feedback and query the LLM for alternative constraint proposals in subsequent rounds.

\paragraph{Termination and fallback.}
The interaction proceeds until a certified result is found or the budget $R$ is exhausted. Individual rounds are not guaranteed to make progress, as LLM proposals may be invalid or repeated. In the latter case, the system invokes the baseline solver $S(F)$ with the remaining time. This provides a fallback to the baseline solver, intended to mitigate coverage regressions, subject to the end-to-end time budget.

\subsection{LLM Interaction}\label{sec:llm}

Each LLM query includes: (1) task specification (IFS or CFS generation); (2) the preprocessed formula with signature information; (3) current focus set for refinement; (4) feedback from prior solver calls.

The LLM returns JSON with $\{\texttt{"IFS"}:[\{t_1\},\ldots]\}$ or $\{\texttt{"CFS"}:[\{c_1\},\ldots]\}$.
We parse each item and reject proposals that (1) use symbols outside the signature of $F'$, (2) fail sort checking, or (3) violate size/format limits; the remaining items are canonicalized and deduplicated. Invalid outputs are dropped. Proposal validity and repetition statistics are reported in Appendix~\ref{app:proposal_stats}. Full prompts are available in the code repository, with output schemas detailed in Appendix~\ref{app:llm-usage}.




\subsection{Correctness Summary}\label{sec:correct}

\textsc{LinguaArray} maintains soundness through certification: the UNSAT branch exploits $F' \Rightarrow F^{\downarrow}_{\mathit{IFS}}$ (Theorem~\ref{thm:unsat}), the SAT branch exploits $F^{\uparrow}_{\mathit{CFS}} \Rightarrow F'$ (Theorem~\ref{thm:sat}). When strengthening yields UNSAT, we avoid spurious UNSAT by explicit implication checking. The LLM does not determine satisfiability and only guides Focus Set construction; all returned results are validated by the solver. To preserve baseline coverage, if guided attempts exhaust their budgets, we invoke S(F) directly.


\section{Experimental Evaluation}
\label{sec:experiments}

We conduct a comprehensive empirical evaluation to assess the effectiveness of \textsc{LinguaArray} for LLM-guided quantified array SMT solving. To facilitate reproducibility, the source code, prompt templates, and benchmark file lists are publicly available at \url{https://github.com/kylelv2000/LinguaArray} under the MIT license.

Our evaluation addresses the following research questions:
\begin{itemize}[left=0em]
    \item \textbf{RQ1:} How effective is \textsc{LinguaArray} compared to state-of-the-art SMT solvers on quantified array formulas?
    \item \textbf{RQ2:} How do different LLMs, including reasoning-enhanced models, affect solving performance?
    \item \textbf{RQ3:} How does the iteration budget impact the overall solving performance?
\end{itemize}

\subsection{Benchmarks}

We evaluate \textsc{LinguaArray} on quantified array formulas from SMT-COMP 2025~\cite{smtcomp2025}, using benchmarks from SMT-LIB, the standard benchmark repository for SMT solvers. We target logics centered on quantified array theories, covering five representative logics: ALIA, ABV, ABVFP, ABVFPLRA, and ANIA. To ensure tractable experimental scale while maintaining representativeness, we apply a sampling strategy: for logics with fewer than 400 instances, we include all available instances; otherwise, we uniformly sample 400 instances (without replacement) using a fixed random seed (42). This yields a total of 1015 instances (ABV: 400, ABVFP: 60, ABVFPLRA: 77, ALIA: 400, ANIA: 78).

\subsection{Experimental Setup}

\paragraph{Hardware and Environment}
All experiments were conducted on a dedicated server equipped with 8 $\times$ Intel(R) Xeon(R) Platinum 8153 CPUs (16 cores per socket, 2 threads per core; 128 cores / 256 hardware threads in total, 2.00 GHz) and 2,048 GB RAM. All experimental runs used identical hardware and software configurations to ensure reproducible and fair comparison.

\paragraph{SMT Solvers}
We employ two state-of-the-art SMT solvers as backend reasoning engines: \textsc{Z3} (v4.15.0)~\cite{DBLP:conf/tacas/MouraB08} and \textsc{CVC5} (v1.3.2)~\cite{DBLP:conf/tacas/BarbosaBBKLMMMN22}. \textsc{Z3} uses its default configuration. For \textsc{CVC5}, we use a tuned baseline selected from commonly used theory options to ensure robust solver performance, and additionally report results for its portfolio-based SMT-COMP 2025 configuration. We provide the full command lines and tuning details in the Appendix for reproducibility.

\paragraph{Large Language Models.}
We evaluate two state-of-the-art LLMs: DeepSeek-V3.2~\cite{liu2025deepseek}, an open-source frontier model, and Gemini-3-Flash-Preview from Google DeepMind~\cite{team2023gemini}. Both models are configured with temperature 0.0 for deterministic responses.

\paragraph{Timeout and Iteration Configuration.}
We use a 1200-second end-to-end timeout per instance, including LLM latency. Unless otherwise specified, we perform 10 iterations, where each iteration executes both the UNSAT and SAT branch.

\subsection{RQ1: Overall Effectiveness}
\begin{table}[ht]
\centering
\footnotesize
\setlength{\tabcolsep}{0.4pt}
\renewcommand{\arraystretch}{0.9}
\caption{Results on five logics with a 1200s timeout (including LLM latency). \textsc{LinguaArray} uses DeepSeek-V3.2-chat. For each logic, we report \#SAT / \#UNSAT and success rate (SR).}
\resizebox{\textwidth}{!}{
\begin{tabular}{lccc ccc ccc ccc ccc | cc}
\toprule
& \multicolumn{3}{c}{ABV (400)}
& \multicolumn{3}{c}{ABVFP (60)}
& \multicolumn{3}{c}{ABVFPLRA (77)}
& \multicolumn{3}{c}{ALIA (400)}
& \multicolumn{3}{c}{ANIA (78)}
& \multicolumn{2}{c}{Overall (1015)} \\
\cmidrule(lr){2-4}\cmidrule(lr){5-7}\cmidrule(lr){8-10}\cmidrule(lr){11-13}\cmidrule(lr){14-16}\cmidrule(lr){17-18}
Method
& SAT & UNSAT & SR
& SAT & UNSAT & SR
& SAT & UNSAT & SR
& SAT & UNSAT & SR
& SAT & UNSAT & SR
& Solved & SR \\
\midrule
\textsc{Z3} (Baseline)
& 34 & 11 & 11.25\%
& 0 & 0 & 0.00\%
& 0 & 0 & 0.00\%
& 51 & 8 & 14.75\%
& 28 & 9 & 47.44\%
& 141 & 13.89\% \\
\textsc{CVC5} (Baseline)
& 93 & 15 & 27.00\%
& 31 & \textbf{3} & 56.67\%
& 45 & 1 & 59.74\%
& 84 & 15 & 24.75\%
& 27 & 7 & 43.59\%
& 321 & 31.63\% \\
\midrule
\textsc{CVC5}-COMP
& 130 & \textbf{82} & 53.00\%
& 37 & \textbf{3} & 66.67\%
& 47 & \textbf{2} & 63.64\%
& 15 & 66 & 20.25\%
& 4 & 2 & 7.69\%
& 388 & 38.23\% \\
\midrule
Harness-only (Z3)
& 34 & 10 & 11.00\%
& 0 & 0 & 0.00\%
& 0 & 0 & 0.00\%
& 52 & 8 & 15.00\%
& 28 & 11 & 50.00\%
& 143 & 14.09\% \\
Randomized (Z3)
& 160 & 19 & 44.75\%
& 0 & 0 & 0.00\%
& 0 & 0 & 0.00\%
& 181 & 17 & 49.50\%
& 45 & 14 & 75.64\%
& 436 & 42.96\% \\
\textsc{LinguaArray(Z3)}
& \textbf{270} & 74 & \textbf{86.00\%}
& 0 & 0 & 0.00\%
& 0 & 0 & 0.00\%
& \textbf{219} & 76 & \textbf{73.75\%}
& \textbf{48} & \textbf{14} & \textbf{79.49\%}
& \textbf{701} & \textbf{69.06\%} \\
\midrule
Harness-only (CVC5)
& 93 & 15 & 27.00\%
& 31 & \textbf{3} & 56.67\%
& 45 & 1 & 59.74\%
& 85 & 15 & 25.00\%
& 27 & 7 & 43.59\%
& 322 & 31.72\% \\
Randomized (CVC5)
& 128 & 23 & 37.75\%
& 33 & \textbf{3} & 60.00\%
& 45 & 1 & 59.74\%
& 149 & 23 & 43.00\%
& 32 & 8 & 51.28\%
& 445 & 43.84\% \\
\textsc{LinguaArray(CVC5)}
& 175 & 72 & 61.75\%
& \textbf{40} & \textbf{3} & \textbf{71.67\%}
& \textbf{53} & 1 & \textbf{70.13\%}
& 186 & \textbf{80} & 66.50\%
& 40 & 9 & 62.82\%
& 659 & 64.93\% \\
\bottomrule
\end{tabular}}
\label{tab:main_results}
\end{table}

Table~\ref{tab:main_results} presents the overall performance comparison across five SMT-LIB logics using DeepSeek as the underlying LLM for \textsc{LinguaArray}. Under the same 1200-second end-to-end timeout, we observed a significant improvement compared to the backend solver. Furthermore, the solved counts increase substantially for both SAT and UNSAT instances, indicating the consistent effectiveness of our two branches. \textsc{LinguaArray(Z3)} achieves the best overall success rate at 69.06\%, solving 701 instances, including 537 SAT instances and 164 UNSAT instances, representing a 397\% improvement over the base \textsc{Z3} solver. Similarly, \textsc{LinguaArray(CVC5)} reaches 64.93\%, improving 105\% over its base solver.

To isolate the contribution of LLM guidance, we evaluate harness-only and structure-aware randomized proposers under the same budget. The harness-only variant uses empty focus sets, while the randomized proposer samples well-typed ground terms for IFS and generates atomic constraints for CFS (see Appendix~\ref{app:nonllm}). The harness-only variants solve 143 and 322 instances, nearly identical to the original backends, while randomized proposers solve 436 and 445 instances. \textsc{LinguaArray} achieves substantially higher coverage, demonstrating the benefit of semantic guidance beyond the search framework itself.

These results significantly surpass the portfolio-based \textsc{CVC5}-COMP at 38.23\%. We further consider virtual-best portfolios, defined as the union of instances solved by different solver configurations. The virtual-best combination of Z3 and CVC5 solves 390 instances (38.4\%), while the union of Z3, \textsc{CVC5}-COMP, and 20 CVC5 configurations reaches 623 (61.4\%). Even compared to this idealized portfolio, \textsc{LinguaArray(Z3)} solves 256 instances outside its coverage.

For ABVFP and ABVFPLRA, \textsc{Z3} produces internal errors when handling the combination of quantifiers, arrays, and floating-point theories, preventing it from solving any instances. The performance of \textsc{LinguaArray(CVC5)} still shows significant improvement compared to its base solver.



\subsection{RQ2: Impact of Different LLMs}
\begin{figure}[ht]
    \centering
    \begin{subfigure}{0.49\linewidth}
        \centering
        \includegraphics[width=\linewidth]{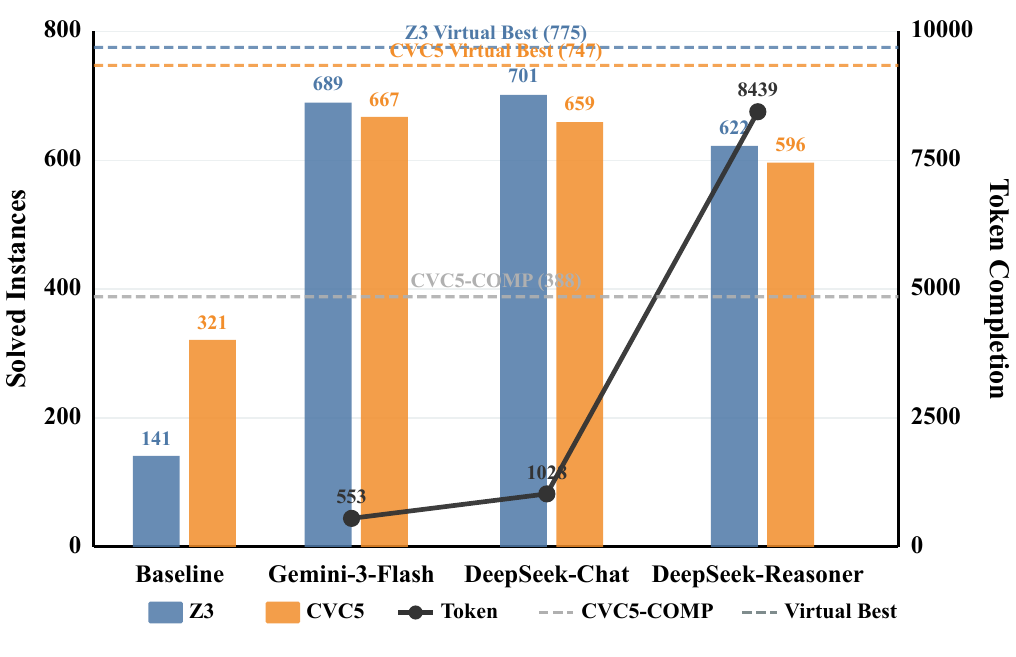}
        \caption{Different LLMs}
        \label{fig:llm_comparison}
        \end{subfigure}
    \hfill
    \begin{subfigure}{0.49\linewidth}
        \centering
        \includegraphics[width=\linewidth]{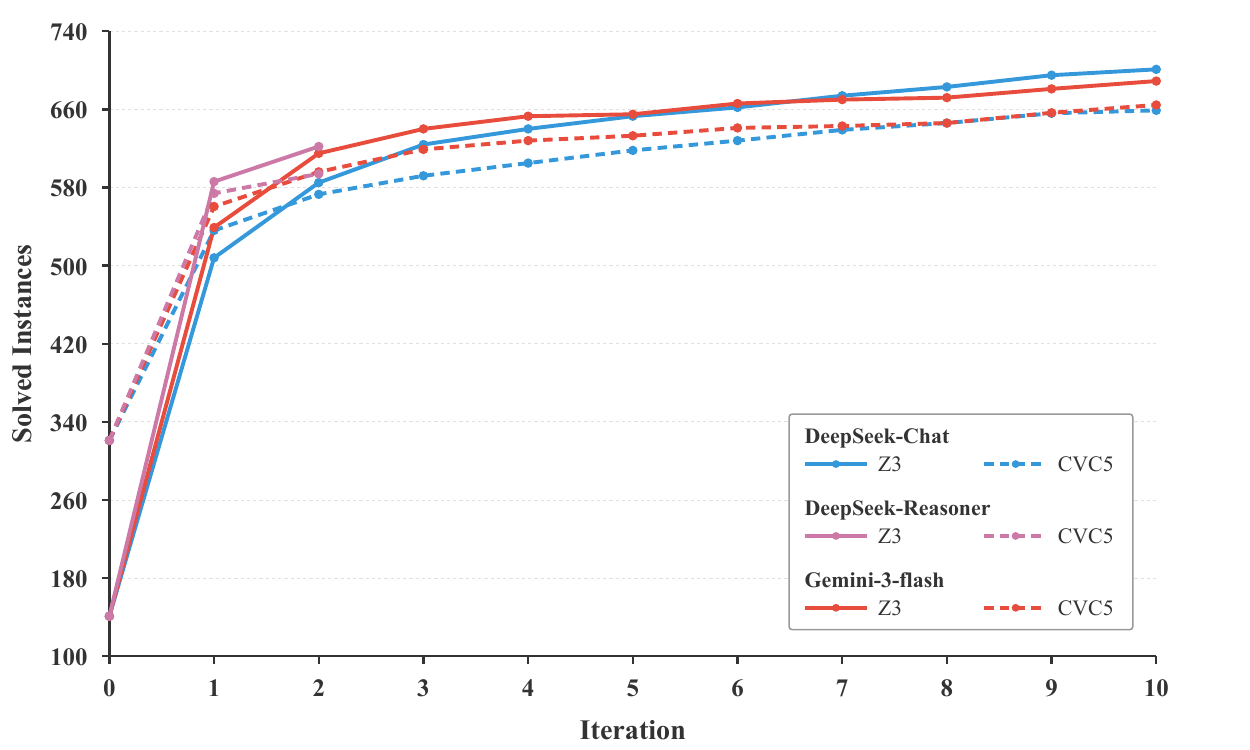}
        \caption{Iteration budget}
        \label{fig:iteration_curve}
    \end{subfigure}
    \caption{Comparison of solved instances across settings: (a) different LLMs; (b) iteration budget under a 1200s timeout. Each point in (b) reports cumulative instances solved within the given budget.}
    \label{fig:main_comparison}
\end{figure}

Figure~\ref{fig:llm_comparison} compares the performance of three LLMs across both base solvers. All three models substantially improve over the baselines, demonstrating that our approach generalizes across different LLMs. However, the results reveal a clear trade-off between reasoning capability and generation efficiency. DeepSeek-V3.2-reasoner produces significantly longer outputs with an average of 8439 tokens per completion due to its explicit reasoning process, resulting in query latencies often approaching 600 seconds. This severely limits the number of iterations possible within the timeout, restricting it to at most two rounds and ultimately leading to lower success rates despite its stronger reasoning capabilities. In contrast, Gemini-3-Flash generates the most concise responses at 553 tokens on average, enabling faster response times and more iterative refinements within the timeout.

The virtual best results demonstrate notable complementarity across LLMs, solving 775 instances for \textsc{Z3} and 747 for \textsc{CVC5}. This represents an additional 74 and 80 instances beyond the best single LLM, respectively, illustrating that different LLMs may use different strategies, thus resulting in different effects on different instances. This complementarity suggests potential for ensemble approaches that combine multiple LLMs.
\subsection{RQ3: Effect of Iteration Budget}

Figure~\ref{fig:iteration_curve} illustrates the relationship between iteration budget and solving capability under the 1200-second timeout constraint. A single iteration already brings substantial improvements, increasing success rates from baseline 13.9\% to 50.1\% for \textsc{Z3} using DeepSeek-Chat. This shows that our approach provides immediate value even without further interaction.

The curves reveal clear diminishing returns. The first iteration contributes the majority of improvements, and each subsequent round progressively adds fewer instances. Most of the gains are achieved within 3 iterations, after which performance generally stabilizes. This pattern indicates that LLMs quickly identify effective strategies in early rounds, with later iterations primarily exploring marginal refinements.

Comparing across LLMs, Gemini-3-Flash achieves stronger first-round performance but slower subsequent growth, while DeepSeek-Chat shows more sustained improvements across iterations. DeepSeek Reasoner (Z3) solved as many as 586 instances in the first round, despite its long processing time, which demonstrates that deeper reasoning can partially compensate for fewer iterations. These observations suggest that the optimal iteration budget depends on the specific LLM's characteristics and the trade-off between response quality and generation efficiency.

\subsection{Discussion}

Our results demonstrate that LLM-guided focal solving substantially improves solving capability across diverse theory combinations. The complementarity observed across different LLMs reveals that diverse reasoning approaches lead to different focal guidance patterns, with different models succeeding on different problem subsets. This suggests potential for ensemble methods to further expand coverage. The diminishing returns pattern indicates that most effective guidance emerges within early iterations, making our approach practical despite LLM query overhead. We further evaluate run-to-run stability using DeepSeek-V4-Flash, the successor to DeepSeek-V3.2. Three independent runs with Z3 solve 689.7 $\pm$ 4.0 instances (mean $\pm$ std), with 602 instances solved in all three runs, indicating stable overall performance despite instance-level variability (Appendix~\ref{app:stability}).

While our method significantly advances performance on complex theory combinations, challenges remain for the deepest theory interactions where formula complexity itself becomes the primary bottleneck, requiring further advances in solving techniques beyond focal guidance. Meanwhile, effectiveness depends on the quality of LLM inference, which may return ineffective proposals and incur unnecessary overhead.

\section{Related Work}

\paragraph{Advanced Quantifier Instantiation.}
Counterexample-guided quantifier instantiation (CEGQI)~\cite{reynolds2015counterexample} and its syntax-guided variant SyQI~\cite{niemetz2021syntax} select instantiation terms by enumerating candidates (optionally from a grammar) and refining them against counterexample models, providing theory-agnostic alternatives to hand-crafted trigger heuristics. QSMA~\cite{bonacina2023qsma} instead computes model-based under- and over-approximations in an abstraction--refinement loop, resembling a lazy approach to quantifier elimination. 
For arrays, the Array Property Fragment~\cite{bradley2006s} yields decidability by restricting quantifiers to arithmetic index guards and value constraints, enabling completeness via finite instantiation; formulas outside this fragment typically require heuristic guidance.

\paragraph{Learning for SMT Solving.}
Jakub\r{u}v et al. use fast gradient-boosted decision tree models over formula features to guide quantifier selection in \textsc{CVC5}, reducing wasted instantiations~\cite{jakubuuv2025machine}.
Beyond quantifier selection, learning has also been applied to instantiation-level decisions: Janota et al. learn to rank candidate ground terms for enumerative instantiation and integrate LightGBM into \textsc{CVC5} to steer instantiation and reduce the number of instantiations~\cite{janota2022towards}.
Similarly, learning-based instance selection inside SMT solvers has been explored by invoking a fast predictor after each instantiation round to decide whether generated instances should be passed to the ground solver, delayed, or discarded~\cite{blanchette2019machine}.
More recently, LLMs have also been initially explored for guiding SMT solving in restricted theories. AquaForte targets quantified formulas over uninterpreted functions by using LLMs to synthesize complete function definitions for solver verification~\cite{lv2026llm}. This generate-and-test approach primarily targets SAT, with limited improvements on UNSAT instances. In contrast, \textsc{LinguaArray} proposes local focus sets rather than complete candidate solutions. Its dual-branch framework uses IFS-based over-approximations for UNSAT reasoning and CFS-based strengthening for SAT reasoning, with additional implication checks to certify UNSAT under strengthening. This enables solver-certified reasoning in both directions.



\section{Conclusion}
We presented \textsc{LinguaArray}, a framework that introduces Semantic Focus Sets as a principled interface between LLM reasoning and SMT solver certification for quantified array formulas. Our two-branch architecture uses Instantiation Focus Sets to guide quantifier instantiation for proving unsatisfiability and Constraint Focus Sets to guide model construction for proving satisfiability, with all results verified by the underlying solver to maintain formal soundness guarantees.

Experimental evaluation on SMT-LIB benchmarks demonstrates substantial improvements, with solve rates advancing from less than 32\% to approximately 70\%, representing 397\% improvement over \textsc{Z3} and 105\% over \textsc{CVC5}. These gains confirm that semantic guidance can effectively identify proof-relevant focal points that conventional heuristics miss. However, this framework can still be further improved because it has limited adaptability to broader and more complex theories, requiring further design. Furthermore, its effectiveness depends on the quality of LLM inference; performance varies depending on the model and problem characteristics.

Future work includes several promising directions. Extending the framework to broader quantified SMT theories beyond arrays, incorporating uninterpreted functions, and developing more unified specifications for focus set generation could expand applicability. Deeper LLM integration, such as using learned refinement strategies or tighter coupling with solver internals, may further improve guidance quality. More broadly, this work demonstrates that semantic reasoning from LLMs can augment symbolic systems while preserving formal guarantees, opening pathways for hybrid approaches in SMT solving.

\begin{ack}
We are grateful to the anonymous reviewers for their comments and suggestions. 
This work was supported by the Strategic Priority Research Program of the Chinese Academy of Sciences, Grant No. XDB1770302 and XDB1770300.
The authors declare no competing interests.
\end{ack}

\bibliographystyle{plainnat}
\bibliography{ijcai26}

@inproceedings{brummayer2008lemmas,
  title={Lemmas on demand for the extensional theory of arrays},
  author={Brummayer, Robert and Biere, Armin},
  booktitle={Proceedings of the Joint Workshops of the 6th International Workshop on Satisfiability Modulo Theories and 1st International Workshop on Bit-Precise Reasoning},
  pages={6--11},
  year={2008}
}

@book{DBLP:series/txtcs/KroeningS16,
  author       = {Daniel Kroening and
                  Ofer Strichman},
  title        = {Decision Procedures - An Algorithmic Point of View, Second Edition},
  series       = {Texts in Theoretical Computer Science. An {EATCS} Series},
  publisher    = {Springer},
  year         = {2016},
  url          = {https://doi.org/10.1007/978-3-662-50497-0},
  doi          = {10.1007/978-3-662-50497-0},
  isbn         = {978-3-662-50496-3},
  bibsource    = {dblp computer science bibliography, https://dblp.org}
}

@incollection{DBLP:series/faia/BarrettSST21,
  author       = {Clark W. Barrett and
                  Roberto Sebastiani and
                  Sanjit A. Seshia and
                  Cesare Tinelli},
  title        = {Satisfiability Modulo Theories},
  booktitle    = {Handbook of Satisfiability - Second Edition},
  series       = {Frontiers in Artificial Intelligence and Applications},
  volume       = {336},
  pages        = {1267--1329},
  publisher    = {{IOS} Press},
  year         = {2021},
  url          = {https://doi.org/10.3233/FAIA201017},
  doi          = {10.3233/FAIA201017},
  bibsource    = {dblp computer science bibliography, https://dblp.org}
}

@inproceedings{DBLP:conf/kbse/CordeiroFM09,
  author       = {Lucas C. Cordeiro and
                  Bernd Fischer and
                  Jo{\~{a}}o Marques{-}Silva},
  title        = {SMT-Based Bounded Model Checking for Embedded {ANSI-C} Software},
  booktitle    = {{ASE} 2009, 24th {IEEE/ACM} International Conference on Automated
                  Software Engineering, Auckland, New Zealand, November 16-20, 2009},
  pages        = {137--148},
  publisher    = {{IEEE} Computer Society},
  year         = {2009},
  url          = {https://doi.org/10.1109/ASE.2009.63},
  doi          = {10.1109/ASE.2009.63},
  bibsource    = {dblp computer science bibliography, https://dblp.org}
}

@article{DBLP:journals/jar/BeyerDW18,
  author       = {Dirk Beyer and
                  Matthias Dangl and
                  Philipp Wendler},
  title        = {A Unifying View on SMT-Based Software Verification},
  journal      = {J. Autom. Reason.},
  volume       = {60},
  number       = {3},
  pages        = {299--335},
  year         = {2018},
  url          = {https://doi.org/10.1007/s10817-017-9432-6},
  doi          = {10.1007/s10817-017-9432-6},
  bibsource    = {dblp computer science bibliography, https://dblp.org}
}

@inproceedings{DBLP:conf/cav/GavrilenkoLFHM19,
  author       = {Natalia Gavrilenko and
                  Hern{\'{a}}n Ponce de Le{\'{o}}n and
                  Florian Furbach and
                  Keijo Heljanko and
                  Roland Meyer},
  editor       = {Isil Dillig and
                  Serdar Tasiran},
  title        = {{BMC} for Weak Memory Models: Relation Analysis for Compact {SMT}
                  Encodings},
  booktitle    = {Computer Aided Verification - 31st International Conference, {CAV}
                  2019, New York City, NY, USA, July 15-18, 2019, Proceedings, Part
                  {I}},
  series       = {Lecture Notes in Computer Science},
  volume       = {11561},
  pages        = {355--365},
  publisher    = {Springer},
  year         = {2019},
  url          = {https://doi.org/10.1007/978-3-030-25540-4\_19},
  doi          = {10.1007/978-3-030-25540-4\_19},
  bibsource    = {dblp computer science bibliography, https://dblp.org}
}

@inproceedings{DBLP:conf/nfm/PeleskaVL11,
  author       = {Jan Peleska and
                  Elena Vorobev and
                  Florian Lapschies},
  editor       = {Mihaela Gheorghiu Bobaru and
                  Klaus Havelund and
                  Gerard J. Holzmann and
                  Rajeev Joshi},
  title        = {Automated Test Case Generation with SMT-Solving and Abstract Interpretation},
  booktitle    = {{NASA} Formal Methods - Third International Symposium, {NFM} 2011,
                  Pasadena, CA, USA, April 18-20, 2011. Proceedings},
  series       = {Lecture Notes in Computer Science},
  volume       = {6617},
  pages        = {298--312},
  publisher    = {Springer},
  year         = {2011},
  url          = {https://doi.org/10.1007/978-3-642-20398-5\_22},
  doi          = {10.1007/978-3-642-20398-5\_22},
  bibsource    = {dblp computer science bibliography, https://dblp.org}
}

@inproceedings{DBLP:conf/tacas/MouraB08,
  author    = {Leonardo Mendon{\c{c}}a de Moura and
               Nikolaj Bj{\o}rner},
  editor    = {C. R. Ramakrishnan and
               Jakob Rehof},
  title     = {{Z3:} An Efficient {SMT} Solver},
  booktitle = {{TACAS} 2008},
  volume    = {4963},
  pages     = {337--340},
  year={2008}
}

@inproceedings{DBLP:conf/tacas/BarbosaBBKLMMMN22,
  author    = {Haniel Barbosa and
               Clark W. Barrett and
               Martin Brain and
               Gereon Kremer and
               Hanna Lachnitt and
               Makai Mann and
               Abdalrhman Mohamed and
               Mudathir Mohamed and
               Aina Niemetz and
               Andres N{\"{o}}tzli and
               Alex Ozdemir and
               Mathias Preiner and
               Andrew Reynolds and
               Ying Sheng and
               Cesare Tinelli and
               Yoni Zohar},
  editor    = {Dana Fisman and
               Grigore Rosu},
  title     = {cvc5: {A} Versatile and Industrial-Strength {SMT} Solver},
  booktitle = {{TACAS} 2022},
  volume    = {13243},
  pages     = {415--442},
  year={2022}
}

@misc{smtcomp2025,
  author = {{SMT-COMP Organizers}},
  title = {{SMT-COMP 2025}},
  howpublished = {\url{https://smt-comp.github.io/2025}},
  year = {2025}
}

@inproceedings{de2007efficient,
  title={Efficient E-matching for SMT solvers},
  author={De Moura, Leonardo and Bj{\o}rner, Nikolaj},
  booktitle={International Conference on Automated Deduction},
  pages={183--198},
  year={2007},
  organization={Springer}
}

@inproceedings{reynolds2018revisiting,
  title={Revisiting enumerative instantiation},
  author={Reynolds, Andrew and Barbosa, Haniel and Fontaine, Pascal},
  booktitle={International Conference on Tools and Algorithms for the Construction and Analysis of Systems},
  pages={112--131},
  year={2018},
  organization={Springer}
}

@inproceedings{becker2019axiom,
  title={The axiom profiler: Understanding and debugging SMT quantifier instantiations},
  author={Becker, Nils and M{\"u}ller, Peter and Summers, Alexander J},
  booktitle={International Conference on Tools and Algorithms for the Construction and Analysis of Systems},
  pages={99--116},
  year={2019},
  organization={Springer}
}

@article{nieuwenhuis2006solving,
  title={Solving SAT and SAT modulo theories: From an abstract Davis--Putnam--Logemann--Loveland procedure to DPLL (T)},
  author={Nieuwenhuis, Robert and Oliveras, Albert and Tinelli, Cesare},
  journal={Journal of the ACM (JACM)},
  volume={53},
  number={6},
  pages={937--977},
  year={2006},
  publisher={ACM New York, NY, USA}
}

@article{vaswani2017attention,
  title={Attention is all you need},
  author={Vaswani, Ashish and Shazeer, Noam and Parmar, Niki and Uszkoreit, Jakob and Jones, Llion and Gomez, Aidan N and Kaiser, {\L}ukasz and Polosukhin, Illia},
  journal={Advances in neural information processing systems},
  volume={30},
  year={2017}
}

@article{wei2022chain,
  title={Chain-of-thought prompting elicits reasoning in large language models},
  author={Wei, Jason and Wang, Xuezhi and Schuurmans, Dale and Bosma, Maarten and Xia, Fei and Chi, Ed and Le, Quoc V and Zhou, Denny and others},
  journal={Advances in neural information processing systems},
  volume={35},
  pages={24824--24837},
  year={2022}
}

@article{lewkowycz2022solving,
  title={Solving quantitative reasoning problems with language models},
  author={Lewkowycz, Aitor and Andreassen, Anders and Dohan, David and Dyer, Ethan and Michalewski, Henryk and Ramasesh, Vinay and Slone, Ambrose and Anil, Cem and Schlag, Imanol and Gutman-Solo, Theo and others},
  journal={Advances in neural information processing systems},
  volume={35},
  pages={3843--3857},
  year={2022}
}

@inproceedings{reynolds2013quantifier,
  title={Quantifier instantiation techniques for finite model finding in SMT},
  author={Reynolds, Andrew and Tinelli, Cesare and Goel, Amit and Krsti{\'c}, Sava and Deters, Morgan and Barrett, Clark},
  booktitle={International Conference on Automated Deduction},
  pages={377--391},
  year={2013},
  organization={Springer}
}

@article{wei2022emergent,
  title={Emergent abilities of large language models},
  author={Wei, Jason and Tay, Yi and Bommasani, Rishi and Raffel, Colin and Zoph, Barret and Borgeaud, Sebastian and Yogatama, Dani and Bosma, Maarten and Zhou, Denny and Metzler, Donald and others},
  journal={arXiv preprint arXiv:2206.07682},
  year={2022}
}

@article{liu2023pre,
  title={Pre-train, prompt, and predict: A systematic survey of prompting methods in natural language processing},
  author={Liu, Pengfei and Yuan, Weizhe and Fu, Jinlan and Jiang, Zhengbao and Hayashi, Hiroaki and Neubig, Graham},
  journal={ACM computing surveys},
  volume={55},
  number={9},
  pages={1--35},
  year={2023},
  publisher={ACM New York, NY}
}

@article{sahoo2024systematic,
  title={A systematic survey of prompt engineering in large language models: Techniques and applications},
  author={Sahoo, Pranab and Singh, Ayush Kumar and Saha, Sriparna and Jain, Vinija and Mondal, Samrat and Chadha, Aman},
  journal={arXiv preprint arXiv:2402.07927},
  year={2024}
}

@inproceedings{ge2009complete,
  title={Complete instantiation for quantified formulas in satisfiability modulo theories},
  author={Ge, Yeting and De Moura, Leonardo},
  booktitle={International Conference on Computer Aided Verification},
  pages={306--320},
  year={2009},
  organization={Springer}
}

@inproceedings{bonacina2023qsma,
  title={QSMA: a new algorithm for quantified satisfiability modulo theory and assignment},
  author={Bonacina, Maria Paola and Graham-Lengrand, St{\'e}phane and Vauthier, Christophe},
  booktitle={International Conference on Automated Deduction},
  pages={78--95},
  year={2023},
  organization={Springer}
}

@inproceedings{reynolds2015counterexample,
  title={Counterexample-guided quantifier instantiation for synthesis in SMT},
  author={Reynolds, Andrew and Deters, Morgan and Kuncak, Viktor and Tinelli, Cesare and Barrett, Clark},
  booktitle={International Conference on Computer Aided Verification},
  pages={198--216},
  year={2015},
  organization={Springer}
}

@article{liu2025deepseek,
  title={Deepseek-v3.2: Pushing the frontier of open large language models},
  author={Liu, Aixin and Mei, Aoxue and Lin, Bangcai and Xue, Bing and Wang, Bingxuan and Xu, Bingzheng and Wu, Bochao and Zhang, Bowei and Lin, Chaofan and Dong, Chen and others},
  journal={arXiv preprint arXiv:2512.02556},
  year={2025}
}

@article{team2023gemini,
  title={Gemini: a family of highly capable multimodal models},
  author={Team, Gemini and Anil, Rohan and Borgeaud, Sebastian and Alayrac, Jean-Baptiste and Yu, Jiahui and Soricut, Radu and Schalkwyk, Johan and Dai, Andrew M and Hauth, Anja and Millican, Katie and others},
  journal={arXiv preprint arXiv:2312.11805},
  year={2023}
}

@inproceedings{niemetz2021syntax,
  title={Syntax-guided quantifier instantiation},
  author={Niemetz, Aina and Preiner, Mathias and Reynolds, Andrew and Barrett, Clark and Tinelli, Cesare},
  booktitle={International Conference on Tools and Algorithms for the Construction and Analysis of Systems},
  pages={145--163},
  year={2021},
  organization={Springer}
}

@inproceedings{bradley2006s,
  title={What’s decidable about arrays?},
  author={Bradley, Aaron R and Manna, Zohar and Sipma, Henny B},
  booktitle={International Workshop on Verification, Model Checking, and Abstract Interpretation},
  pages={427--442},
  year={2006},
  organization={Springer}
}

@article{jakubuuv2025machine,
  title={Machine learning for quantifier selection in cvc5},
  author={Jakub{\r{u}}v, Jan and Janota, Mikol{\'a}{\v{s}} and Piepenbrock, Jelle and Urban, Josef},
  journal={International Journal of Approximate Reasoning},
  pages={109602},
  year={2025},
  publisher={Elsevier}
}

@inproceedings{DBLP:conf/lics/StumpBDL01,
  author       = {Aaron Stump and
                  Clark W. Barrett and
                  David L. Dill and
                  Jeremy R. Levitt},
  title        = {A Decision Procedure for an Extensional Theory of Arrays},
  booktitle    = {16th Annual {IEEE} Symposium on Logic in Computer Science, Boston,
                  Massachusetts, USA, June 16-19, 2001, Proceedings},
  pages        = {29--37},
  publisher    = {{IEEE} Computer Society},
  year         = {2001},
  url          = {https://doi.org/10.1109/LICS.2001.932480},
  doi          = {10.1109/LICS.2001.932480},
  bibsource    = {dblp computer science bibliography, https://dblp.org}
}

@article{liu2024deepseek,
  title={Deepseek-v3 technical report},
  author={Liu, Aixin and Feng, Bei and Xue, Bing and Wang, Bingxuan and Wu, Bochao and Lu, Chengda and Zhao, Chenggang and Deng, Chengqi and Zhang, Chenyu and Ruan, Chong and others},
  journal={arXiv preprint arXiv:2412.19437},
  year={2024}
}

@inproceedings{kong2024better,
  title={Better zero-shot reasoning with role-play prompting},
  author={Kong, Aobo and Zhao, Shiwan and Chen, Hao and Li, Qicheng and Qin, Yong and Sun, Ruiqi and Zhou, Xin and Wang, Enzhi and Dong, Xiaohang},
  booktitle={Proceedings of the 2024 Conference of the North American Chapter of the Association for Computational Linguistics: Human Language Technologies (Volume 1: Long Papers)},
  pages={4099--4113},
  year={2024}
}

@inproceedings{janota2022towards,
  title={Towards learning quantifier instantiation in SMT},
  author={Janota, Mikol{\'a}{\v{s}} and Piepenbrock, Jelle and Piotrowski, Bartosz},
  booktitle={25th International Conference on Theory and Applications of Satisfiability Testing (SAT 2022)},
  pages={7--1},
  year={2022},
  organization={Schloss Dagstuhl--Leibniz-Zentrum f{\"u}r Informatik}
}

@inproceedings{blanchette2019machine,
  title={Machine learning for instance selection in SMT solving},
  author={Blanchette, Jasmin Christian and El Ouraoui, Daniel and Fontaine, Pascal and Kaliszyk, Cezary},
  booktitle={AITP 2019-4th Conference on Artificial Intelligence and Theorem Proving},
  year={2019}
}

@inproceedings{lv2026llm,
  title={LLM-guided quantified SMT solving over uninterpreted functions},
  author={Lv, Kunhang and Dong, Yuhang and Han, Rui and Jia, Fuqi and Ma, Feifei and Zhang, Jian},
  booktitle={Proceedings of the AAAI Conference on Artificial Intelligence},
  volume={40},
  number={17},
  pages={14304--14312},
  year={2026}
}

\clearpage
\appendix

\section{Detailed Algorithm}
\label{app:algorithm}

Algorithm~\ref{alg:main} gives the full procedure of \textsc{LinguaArray}.
The algorithm follows the dual-branch reasoning strategy described in Section~\ref{sec:reasoning}: the UNSAT branch constructs an Instantiation Focus Set (IFS) to obtain a finite instantiation-based approximation, while the SAT branch constructs a Constraint Focus Set (CFS) to guide model search through focused strengthening. The feedback state $\mathit{fb}$ records information from unsuccessful solver calls and is used to refine subsequent LLM proposals.

\begin{algorithm}[ht]
\caption{\textsc{LinguaArray}: Semantic Focus Set Guided Solving}
\label{alg:main}
\begin{algorithmic}[1]
\REQUIRE Formula $F$, solver $\mathcal{S}$, budget $R$
\ENSURE $\mathsf{sat}$, $\mathsf{unsat}$, or $\mathsf{unknown}$
\STATE $(F', \Sigma) \leftarrow \textsc{Preprocess}(F)$; $\mathit{fb} \leftarrow \emptyset$
\FOR{$r = 1$ to $R$}
    \STATE \COMMENT{UNSAT Branch: Instantiation Focus Set}
    \STATE $\mathit{IFS} \leftarrow \textsc{ProposeIFS}(F', \mathit{fb})$
    \STATE $\mathit{res}_1 \leftarrow \mathcal{S}(F^{\downarrow}_{\mathit{IFS}})$
    \IF{$\mathit{res}_1 = \mathsf{unsat}$} \RETURN $\mathsf{unsat}$ \ENDIF
    \IF{$\mathit{res}_1 = \mathsf{sat}$} \STATE $\mathit{fb} \leftarrow \mathit{fb} \cup \textsc{ExtractFeedback}(\mathit{res}_1)$ \ENDIF
    \STATE \COMMENT{SAT Branch: Constraint Focus Set}
    \STATE $\mathit{CFS} \leftarrow \textsc{ProposeCFS}(F', \mathit{fb})$

    \STATE $\mathit{res}_2 \leftarrow \mathcal{S}(F^{\uparrow}_{\mathit{CFS}})$
    \IF{$\mathit{res}_2 = \mathsf{sat}$} \RETURN $\mathsf{sat}$ \ENDIF
    \IF{$\mathit{res}_2 = \mathsf{unsat}$ AND $\textsc{Certify}(F', \mathit{CFS})$} \RETURN $\mathsf{unsat}$ \ENDIF
    \IF{$\mathit{res}_2 = \mathsf{unsat}$} \STATE $\mathit{fb} \leftarrow \mathit{fb} \cup \textsc{ExtractFeedback}(\mathit{res}_2)$ \ENDIF
\ENDFOR
\RETURN $\mathcal{S}(F)$ \COMMENT{Fallback with remaining timeout}
\end{algorithmic}
\end{algorithm}

\section{SMT Solver Backends and Configurations}
\label{app:solver-config}

We provide the complete solver configurations used in our experiments.
We use two state-of-the-art SMT solvers as backend reasoning engines: 
\textsc{Z3} (v4.15.0)~\footnote{\url{https://github.com/Z3Prover/z3}} and \textsc{CVC5} (v1.3.2)~\footnote{\url{https://github.com/cvc5/cvc5}}. Both solvers represent the current state of the art in SMT solving and have demonstrated strong performance across diverse logics.

\paragraph{Z3 Configuration.}
For Z3, we use the default configuration, which has been tuned for general-purpose SMT solving across various theories.
The command line is:

\begin{quote}
z3 input.smt2
\end{quote}

\paragraph{\textsc{CVC5} Baseline Configuration.}
\textsc{CVC5}'s performance can depend substantially on theory-specific option choices, especially for benchmarks involving quantified formulas and rich theory reasoning. To ensure a robust baseline solver, we select a tuned baseline configuration from commonly used options for quantified formulas and theory reasoning.
The command line is:

\begin{quote}
cvc5 --finite-model-find --arrays-exp --fp-exp --mbqi --enum-inst --no-cegqi --no-sygus-inst input.smt2
\end{quote}
These options enable model-based quantifier instantiation, finite model finding, enumerative instantiation, and experimental support for array and floating-point theories. The complete results covering all combinations can be found in Appendix~\ref{app:cvc5-param}.

\paragraph{\textsc{CVC5} SMT-COMP Configuration.}
For comparison, we also evaluate the portfolio-based configuration used in SMT-COMP2025~\footnote{\url{https://smt-comp.github.io/2025}}. This configuration represents \textsc{CVC5}'s competition-winning setup and \textsc{CVC5}'s best competitive performance and is included as a strong reference point for comparison.
The command line is:

\begin{quote}
cvc5 -L smt2.6 --no-incremental --no-type-checking --no-interactive --fp-exp --use-portfolio input.smt2
\end{quote}

The \texttt{--use-portfolio} option enables \textsc{CVC5}'s portfolio mode, which sequentially runs multiple internal solving strategies with assigned time budgets.
This allows \textsc{CVC5} to benefit from different configurations within a single solver invocation and often improves robustness across heterogeneous SMT instances.

\section{\textsc{CVC5} Parameter Study}
\label{app:cvc5-param}

We report an auxiliary parameter study for \textsc{CVC5} on our evaluation set.
All runs use a uniform timeout of 1200 seconds per instance.
We counted the total number of solutions and the number of SATs and UNSATs for each parameter combination.

\subsection{Flag glossary}
\label{app:cvc5-glossary}
\begin{table}[!h]
  \centering
  \caption{Glossary of components and the corresponding \textsc{CVC5} flags.}
  \begin{tabular}{ll}
    \toprule
    Term & \textsc{CVC5} parameters \\
    \midrule
    fmf & \texttt{--finite-model-find} \\
    arr  & \texttt{--arrays-exp} \\
    fp & \texttt{--fp-exp} \\
    mbqi & \texttt{--mbqi} \\
    enum & \texttt{--enum-inst} \\
    cegqi & \texttt{--cegqi}\\
    sygus & \texttt{--sygus-inst}\\
    \bottomrule
  \end{tabular}
  \label{tab:cvc5-glossary}
\end{table}

\subsection{Base setting (fmf/arr/fp)}
\label{app:cvc5-base}

We first evaluate the effect of enabling the three base parameters fmf, arr and fp.
Table~\ref{tab:cvc5-base} shows that enabling all three achieves the best overall solved count.
Hence, we use the \textbf{fmf\_arr\_fp} backbone as the default setting for subsequent experiments.

\begin{table}[ht]
  \centering
  \caption{Effect of base parameters.}
  \begin{tabular}{lrrr}
    \toprule
    Configuration & SAT & UNSAT & Total \\
    \midrule
    baseline         & 36 & 24 & 60 \\
    fmf              & 50 & 15 & 65 \\
    fmf\_arr         & 51 & 15 & 66 \\
    fmf\_arr\_fp     & 51 & 15 & 66 \\
    \bottomrule
  \end{tabular}
  \label{tab:cvc5-base}
\end{table}

\subsection{Component ablation}
\label{app:cvc5-ablation}

Next, we ablate four components: mbqi, enum, cegqi, and sygus instantiation.
Configuration names list enabled components; disabled components are turned off using the corresponding \texttt{--no-...} flags as specified in Table~\ref{tab:cvc5-glossary}.
Table~\ref{tab:cvc5-combos} summarizes all 16 combinations.
The best-performing configuration is \textbf{fmf\_arr\_fp\_mbqi\_enum}, which attains the largest solved count.

\begin{table}[ht]
  \centering
  \caption{Ablation results.}
  \begin{tabular}{lrrr}
    \toprule
    Configuration & SAT & UNSAT & Total \\
    \midrule
    \textbf{fmf\_arr\_fp\_mbqi\_enum}                     & 280 &  41 & \textbf{321} \\
    fmf\_arr\_fp\_mbqi                                    & 271 &  32 & 303 \\
    fmf\_arr\_fp\_mbqi\_enum\_cegqi                       & 260 &  39 & 299 \\
    fmf\_arr\_fp\_mbqi\_cegqi                             & 249 &  33 & 282 \\
    fmf\_arr\_fp\_enum\_cegqi                             &  53 & 136 & 189 \\
    fmf\_arr\_fp\_enum\_cegqi\_sygus                      & 137 &  49 & 186 \\
    fmf\_arr\_fp\_cegqi\_sygus                            & 137 &  49 & 186 \\
    fmf\_arr\_fp\_enum\_sygus                             & 131 &  50 & 181 \\
    fmf\_arr\_fp\_sygus                                   & 131 &  50 & 181 \\
    fmf\_arr\_fp\_enum                                    &  45 & 134 & 179 \\
    fmf\_arr\_fp\_mbqi\_cegqi\_sygus                       & 118 &  50 & 168 \\
    fmf\_arr\_fp\_mbqi\_enum\_cegqi\_sygus                 & 118 &  49 & 167 \\
    fmf\_arr\_fp\_mbqi\_enum\_sygus                        & 112 &  48 & 160 \\
    fmf\_arr\_fp\_mbqi\_sygus                              & 112 &  48 & 160 \\
    fmf\_arr\_fp\_cegqi                                    &  51 &  15 &  66 \\
    fmf\_arr\_fp                                           &  43 &  15 &  58 \\
    \bottomrule
  \end{tabular}
  \label{tab:cvc5-combos}
\end{table}

\section{Runtime Analysis}
\label{app:runtime}

In addition to the solved-instance counts reported in Table~\ref{tab:main_results}, we also report the corresponding runtime statistics in Table~\ref{tab:appendix-runtime}. For unsolved instances, we count the runtime as the timeout value of 1200 seconds.

\begin{table}[ht]
\centering
\caption{Runtime statistics corresponding to the results in Table~\ref{tab:main_results}. Unsolved instances are counted as 1200 seconds.}
\label{tab:appendix-runtime}
\begin{tabular}{lccc}
\toprule
Method & Avg.\ total time & \#Solved & Avg.\ time (solved only) \\
\midrule
\textsc{Z3} (Baseline) & 1033.41s & 141 & 0.79s \\
\textsc{CVC5} (Baseline) & 820.75s & 321 & 0.80s \\
\textsc{LinguaArray(Z3)} & 424.15s & 701 & 76.63s \\
\textsc{LinguaArray(CVC5)} & 469.91s & 659 & 75.51s \\
\bottomrule
\end{tabular}
\end{table}

Most baseline-solved instances finish within only a few seconds; otherwise they often run until the 1200s timeout.
\textsc{LinguaArray} adds extra cost from LLM calls and repeated solver checks. This is why its average time on solved instances is higher than that of the baseline solvers. However, \textsc{LinguaArray} substantially increases the number of instances solved within the same end-to-end budget. This shows that the extra LLM-guided reasoning cost is very useful when it helps the solver avoid long unsuccessful searches.

\subsection{Run-to-Run Stability}
\label{app:stability}

Since the original DeepSeek-V3.2 model is no longer available through its official API, we conduct three independent runs using its successor, DeepSeek-V4-Flash, with Z3 as the backend. All runs use non-reasoning mode, temperature 0, 10 refinement rounds, and a 1200-second timeout on the same 1015 instances.

The three runs solve 685, 692, and 692 instances, respectively, yielding a mean of 689.7 and a sample standard deviation of 4.0 (67.95\% success rate). Among them, 602 instances are solved in all three runs, while 762 are solved in at least one run.

For the 602 consistently solved instances, the average end-to-end runtime is 44.6 seconds. The median per-instance runtime standard deviation is 18.2 seconds, with a median coefficient of variation of 61\%. These results indicate stable overall solved counts, although individual instances exhibit variability in both outcomes and runtime.

\section{Non-LLM Proposer Baselines}
\label{app:nonllm}

We evaluate two non-LLM baselines to isolate the contribution of LLM guidance. The harness-only baseline uses empty focus sets while retaining preprocessing, dual-branch execution, validation, and fallback.

The randomized heuristic proposer first extracts all well-typed ground subterms from $F'$, including nested \texttt{select}/\texttt{store} terms. These terms are grouped by sort, deduplicated, ordered by term size, and supplemented with basic theory-specific constants. For IFS, witness terms are uniformly sampled from the corresponding sort-specific pools for each bound variable. For CFS, well-typed atomic constraints, including equalities, disequalities, and order relations, are randomly constructed, with trivial or invalid constraints filtered out.

Each round independently resamples proposals, while the validation, certification, refinement, and fallback mechanisms remain identical to \textsc{LinguaArray}. All experiments use the same 1200-second budget.

\section{LLM Details}
\label{app:llm-usage}

This appendix provides the implementation details needed to reproduce the LLM interface used by \textsc{LinguaArray}: prompt templates, output schemas, response parsing, proposal validation, and cross-round feedback.

\subsection{Prompt Format}
\label{app:llm-prompts}

Each LLM request contains a short shared system instruction and one branch-specific user message. The system instruction asks the model to act as an SMT and formal verification assistant and to return only a JSON object. The user message contains the preprocessed formula $F'$, the branch objective, and the required output schema.

We use different JSON keys for the two branches. The SAT branch returns \texttt{CFS}; the UNSAT branch returns \texttt{IFS}.

\paragraph{SAT branch.}
The SAT branch asks for a small set of ground Boolean constraints over the signature of $F'$. Each constraint must be directly usable inside an SMT-LIB \texttt{assert}. The expected response format is:

\begin{lstlisting}[
    basicstyle=\ttfamily,
    breaklines=true,
    columns=fullflexible,
    xleftmargin=2em,
    framexleftmargin=1em
]
{
  "satisfiability": "sat" | "unsat" | "unknown",
  "chain_of_thought": "<brief explanation>",
  "constraints": [
    {
      "name": "<short label>",
      "constraint": "<SMT-LIB Boolean expression>",
      "type": "assignment" | "disequality" | "...",
      "confidence": <float in [0, 1]>
    }
  ]
}
\end{lstlisting}

The prompt forbids new symbols, bound variables, and wrapping the constraint with \texttt{assert}. It also asks the model to prefer small and concrete constraints, such as equalities, disequalities, bounds, and array-cell constraints.

\paragraph{UNSAT branch.}
The UNSAT branch asks for witness terms for eligible quantified variables. The prompt lists the extracted quantifiers and the top-level declared symbols that may be used to build witness terms. The expected response format is:
\begin{lstlisting}[
    basicstyle=\ttfamily,
    breaklines=true,
    columns=fullflexible,
    xleftmargin=2em,
    framexleftmargin=1em
]
{
  "chain_of_thought": "<brief explanation>",
  "instantiations": [
    {
      "quantifier_id": <id from prompt>,
      "witness_map": {
        "<bound variable>": "<SMT term>"
      },
      "confidence": <float in [0,1]>
    }
  ],
  "unsat_prediction": "high" | "medium" | "low"
}
\end{lstlisting}

Witness terms may use constants, declared symbols, arithmetic or bit-vector operators, and array operations such as \texttt{select} and \texttt{store}. They are not accepted unless they pass the validation checks in Section~\ref{app:validation}.

The complete prompt templates are available in the code repository. The implementation uses only the parsed JSON fields above. Although the prompt asks for a short explanation field and satisfiability, these fields are used only for debugging and auditability. The solver never treats the explanation as a proof. Only parsed, well-typed SMT terms and constraints are passed to the backend solver.

\subsection{Response Parsing}
\label{app:parsing}

The parser expects a single JSON object. If direct parsing fails, it applies the following recovery steps in order:

\begin{enumerate}
\item remove text enclosed by \texttt{<think>}...\texttt{</think>} or \texttt{<thinking>}...\texttt{</thinking>} tags;
\item extract the first JSON fenced block, or otherwise the first unannotated fenced block;
\item remove simple trailing commas and repair unambiguous single-quoted keys;
\item if the output appears truncated, search for the longest parseable JSON prefix.
\end{enumerate}

If all recovery steps fail, the response is treated as empty for that round. We do not issue an additional LLM call only to repair formatting errors.

\subsection{Proposal Validation}
\label{app:validation}

Parsed proposals are checked before they are used. Invalid items are discarded rather than rewritten.

\paragraph{Constraint proposals.}
A CFS item is accepted only if its \texttt{constraint} field parses as a Boolean SMT-LIB expression, uses only symbols declared in $F'$, and does not refer to bound variables. Accepted constraints are canonicalized, deduplicated, and mapped back through $\Sigma^{-1}$ before being passed to the solver.

\paragraph{Instantiation proposals.}
An IFS item is accepted only if the referenced quantifier id exists and the \texttt{witness\_map} assigns one well-typed SMT term to every bound variable of that quantifier. The witness terms are parsed in the same context as $F'$ and checked against the declared variable sorts.

\subsection{Cross-Round Feedback}
\label{app:iteration}

If a round does not produce a certified answer, the failed proposals are added to the feedback state for later rounds. The two branches keep separate feedback records: SAT rounds receive previous constraints, while UNSAT rounds receive previous instantiations.

The feedback block has the following form:

\begin{lstlisting}[
  basicstyle=\ttfamily,
  breaklines=true,
  columns=fullflexible
]
## Previous Attempts

Previous constraints:    (SAT branch only)
  - <constraint string 1>
  - <constraint string 2>
  ...

Previous instantiations: (UNSAT branch only)
  - <witness map 1>
  - <witness map 2>
  ...

Consider: refining these, adding more constraints, or trying different approaches.
\end{lstlisting}

The feedback is used only as prompt context. It is not interpreted as evidence for either satisfiability or unsatisfiability.

\subsection{Proposal Statistics}
\label{app:proposal_stats}

We analyze the validity and cross-round repetition of LLM proposals. For DeepSeek-V3.2, 1.65\% of CFS constraints are invalid due to parsing, type, or symbol errors. For IFS, 13.2\% are invalid, including 12.5\% with parsing or type errors and 0.7\% with missing bound variables. The corresponding invalid rates for Gemini are 3.7\% for CFS and 10.2\% for IFS.

In later rounds, 59.5\% of CFS items and 65.1\% of IFS items have appeared in previous rounds. Such repetitions may still be useful, as solving can require combinations of multiple constraints or instantiations.

\newpage
\section*{NeurIPS Paper Checklist}

\begin{enumerate}

\item {\bf Claims}
    \item[] Question: Do the main claims made in the abstract and introduction accurately reflect the paper's contributions and scope?
    \item[] Answer: \answerYes{} 
    \item[] Justification: The abstract and introduction mentioned three core contributions.

\item {\bf Limitations}
    \item[] Question: Does the paper discuss the limitations of the work performed by the authors?
    \item[] Answer: \answerYes{} 
    \item[] Justification: The limitations are written in the Discussion and Conclusion section.

\item {\bf Theory assumptions and proofs}
    \item[] Question: For each theoretical result, does the paper provide the full set of assumptions and a complete (and correct) proof?
    \item[] Answer: \answerYes{} 
    \item[] Justification: Both Theorem 1 and Theorem 2 have proof.

    \item {\bf Experimental result reproducibility}
    \item[] Question: Does the paper fully disclose all the information needed to reproduce the main experimental results of the paper to the extent that it affects the main claims and/or conclusions of the paper (regardless of whether the code and data are provided or not)?
    \item[] Answer: \answerYes{} 
    \item[] Justification: The detailed configuration has been provided, and the complete code is available in the supplement.

\item {\bf Open access to data and code}
    \item[] Question: Does the paper provide open access to the data and code, with sufficient instructions to faithfully reproduce the main experimental results, as described in supplemental material?
    \item[] Answer: \answerYes{} 
    \item[] Justification: The complete code is available in the supplement.

\item {\bf Experimental setting/details}
    \item[] Question: Does the paper specify all the training and test details (e.g., data splits, hyperparameters, how they were chosen, type of optimizer) necessary to understand the results?
    \item[] Answer: \answerYes{} 
    \item[] Justification: Section 4 provides a detailed explanation of these. And no model training is performed; the LLMs are used as fixed black-box proposal generators.

\item {\bf Experiment statistical significance}
    \item[] Question: Does the paper report error bars suitably and correctly defined or other appropriate information about the statistical significance of the experiments?
    \item[] Answer: \answerNo{} 
    \item[] Justification: We acknowledge that LLM-based evaluation may introduce some variability. To minimize this effect and improve reproducibility, we use deterministic decoding by setting the temperature to 0 for all LLM calls. In our experiments, the observed gains are large and consistent enough that small fluctuations caused by evaluation variability are unlikely to change the main conclusions. Although we do not provide error bars, confidence intervals, or formal statistical significance tests, a detailed statistical uncertainty analysis is not the primary focus of this work and could be further explored in future work.

\item {\bf Experiments compute resources}
    \item[] Question: For each experiment, does the paper provide sufficient information on the computer resources (type of compute workers, memory, time of execution) needed to reproduce the experiments?
    \item[] Answer: \answerYes{} 
    \item[] Justification: Section 4 provides a detailed description of these.
    
\item {\bf Code of ethics}
    \item[] Question: Does the research conducted in the paper conform, in every respect, with the NeurIPS Code of Ethics \url{https://neurips.cc/public/EthicsGuidelines}?
    \item[] Answer: \answerYes{} 
    \item[] Justification: The research conforms to the NeurIPS Code of Ethics.

\item {\bf Broader impacts}
    \item[] Question: Does the paper discuss both potential positive societal impacts and negative societal impacts of the work performed?
    \item[] Answer: \answerNo{} 
    \item[] Justification: The paper discusses positive application domains of improved SMT solving, such as software verification, program analysis, model checking, and automated test generation, but it does not include a dedicated discussion of negative societal impacts. We believe the work is primarily foundational and does not introduce a direct human-facing decision system, personal-data model, or generative content system; potential negative impacts are indirect and mainly relate to possible use of stronger verification or program-analysis tools in dual-use software contexts.
    
\item {\bf Safeguards}
    \item[] Question: Does the paper describe safeguards that have been put in place for responsible release of data or models that have a high risk for misuse (e.g., pre-trained language models, image generators, or scraped datasets)?
    \item[] Answer: \answerNA{} 
    \item[] Justification: This paper poses no such risks.

\item {\bf Licenses for existing assets}
    \item[] Question: Are the creators or original owners of assets (e.g., code, data, models), used in the paper, properly credited and are the license and terms of use explicitly mentioned and properly respected?
    \item[] Answer: \answerYes{} 
    \item[] Justification: The paper credits the existing assets used in the experiments, including SMT-LIB / SMT-COMP benchmarks, Z3, CVC5, and the LLMs used for proposal generation.

\item {\bf New assets}
    \item[] Question: Are new assets introduced in the paper well documented and is the documentation provided alongside the assets?
    \item[] Answer: \answerYes{} 
    \item[] Justification: The complete code is available in the supplement.

\item {\bf Crowdsourcing and research with human subjects}
    \item[] Question: For crowdsourcing experiments and research with human subjects, does the paper include the full text of instructions given to participants and screenshots, if applicable, as well as details about compensation (if any)? 
    \item[] Answer: \answerNA{} 
    \item[] Justification: This paper does not involve crowdsourcing nor research with human subjects.

\item {\bf Institutional review board (IRB) approvals or equivalent for research with human subjects}
    \item[] Question: Does the paper describe potential risks incurred by study participants, whether such risks were disclosed to the subjects, and whether Institutional Review Board (IRB) approvals (or an equivalent approval/review based on the requirements of your country or institution) were obtained?
    \item[] Answer: \answerNA{} 
    \item[] Justification: This paper does not involve crowdsourcing nor research with human subjects.

\item {\bf Declaration of LLM usage}
    \item[] Question: Does the paper describe the usage of LLMs if it is an important, original, or non-standard component of the core methods in this research? Note that if the LLM is used only for writing, editing, or formatting purposes and does \emph{not} impact the core methodology, scientific rigor, or originality of the research, declaration is not required.
    \item[] Answer: \answerYes{} 
    \item[] Justification: LLMs are a core component of \textsc{LinguaArray}: they generate Instantiation Focus Sets and Constraint Focus Sets used to guide SMT solving. The paper describes this usage in the methodology and evaluates different LLMs experimentally.

\end{enumerate}

\end{document}